%% file: main.tex
\documentclass[letterpaper, 10 pt, conference]{ieeeconf}
\usepackage{amsmath}
\usepackage{amssymb}

\usepackage{tabularx}
\usepackage{graphicx}
\usepackage{bm}
\usepackage{color}
\usepackage{float}
\usepackage{units}
\usepackage{url}
\usepackage{hyperref}
\usepackage{balance}
\usepackage{amsmath}

\makeatletter 
\def\endfigure{\end@float} 
\def\endtable{\end@float}
\makeatother

\newcommand{\pos}{\bm{p}}

\usepackage{subcaption}
\usepackage{caption}

\renewcommand{\unit}[1]{{\rm #1} }

\newcommand{\update}[1]{\textcolor{black}{#1}}

\IEEEoverridecommandlockouts

\begin{document} 

% Paper title (needs to change)
\title{\Large \bf 			
Force-and-moment-based Model Predictive Control for Achieving Highly Dynamic Locomotion on Bipedal Robots
}

\author{Junheng Li and Quan Nguyen\thanks{Junheng Li and Quan Nguyen are with the Department of Aerospace and Mechanical Engineering, University of Southern California, Los Angeles, CA 90089.
email:{\tt\small junhengl@usc.edu, quann@usc.edu}}%
}%													
	
% make the title area
\maketitle
%Pins
%Junheng Li 150116

\begin{abstract}

In this paper, we propose a novel framework on force-and-moment-based Model Predictive Control (MPC) for dynamic legged robots. Specifically, we present a formulation of MPC designed for 10 degree-of-freedom (DoF) bipedal robots using simplified rigid body dynamics with input forces and moments. This MPC controller will calculate the optimal inputs applied to the robot, including 3-D forces and 2-D moments at each foot. These desired inputs will then be generated by mapping these forces and moments to motor torques of 5 actuators on each leg. We evaluate our proposed control design on physical simulation of a 10 degree-of-freedom (DoF) bipedal robot. The robot can achieve fast walking speed up to $ 1.6\:\unit{m/s}$ on rough terrain, with accurate velocity tracking. With the same control framework, our proposed approach can achieve a wide range of dynamic motions including walking, hopping, and running using the same set of control parameters.

\end{abstract}

% Introduction and literature review
\input{Sections/Introduction.tex}
% Robot Model and Simulation
\input{Sections/RobotModel.tex}
% Main theory 
\input{Sections/DynamicsAndControl.tex}
% results
\input{Sections/SImulationResults.tex}
% Conclusion
\input{Sections/Conclusion.tex}

\balance
\bibliographystyle{ieeetr}
\bibliography{reference}

% Document end
\end{document}

%% file: Sections/Introduction.tex
%!TEX root = ../Main.tex

\section{Introduction}
\label{sec:Introduction}

The motivation of studying bipedal robots is widely promoted by commercial and sociological interests \cite{westervelt2018feedback}. The desired outcomes of bipedal robot applications range from replacing humans in hazardous operations \cite{chen2020a}, in which it requires highly dynamic robots in unknown complex terrains, to the development of highly functional bipedal robot applications in the medical field and rehabilitation processes such as recent development in research of powered lower-limb prostheses for the disabled \cite{zhao2017first}. 

%Building effective force-based controller remains of the most important cornerstones of achieving dynamic motions while keeping balance for legged robots, because bipedal robot has inherently less stable dynamical system, compared to other sub-classes of legged robots such as quadruped robots 

% \update{Building an effective force-based controller remains one of the most important cornerstones of achieving dynamic motions while keeping balance for legged robots, because bipedal robots are inherently less stable dynamical systems, compared to other sub-classes of legged robots such as quadruped robots} \cite{westervelt2018feedback}. 

%There are many controller design and choices that are employed in bipedal balance control, including controllers that utilize 
\update{There are many control strategies that can be used for control of bipedal robots, such as} Zero-Moment-Point (ZMP) or using spring-loaded inverted pendulum (SLIP) model \cite{ames2012dynamically,kajita2006biped,holmes2006dynamics}. Both methods have had success in maintaining stable locomotion of bipedal robots (e.g. \cite{ames2012dynamically,holmes2006dynamics}). Hybrid Zero Dynamics method is another control framework utilizes input-output linearization, a non-linear feedback controller, with virtual constraints that allows dynamic walking on under-actuated bipedal robots \cite{westervelt2018feedback, nguyen2016dynamic, nguyen2017dynamic}. 
% (This is not published nor open source. So, we should not mention it here)Others use forced-based control such as Model Predictive Control (MPC) to maintain stability during dynamically motions \cite{levineblackbird}. 
Recently, force-based MPC control was introduced for dynamic quadruped robots \cite{di2018dynamic}, allowing the robot to perform a wide range of dynamic gaits with robustness to rough terrains. One advantage of the MPC framework in dynamic locomotion is that the controller can predict future motions that may cause instability due to under-actuation and stabilize the system by solving for optimal inputs based on the prediction. 
% There are many successful controller designs that employs MPC control during dynamical motions in the world of legged robots (e.g. \cite{levineblackbird,di2018dynamic}). 

\begin{figure}[t]
		\center
		\includegraphics[width=1 \columnwidth]{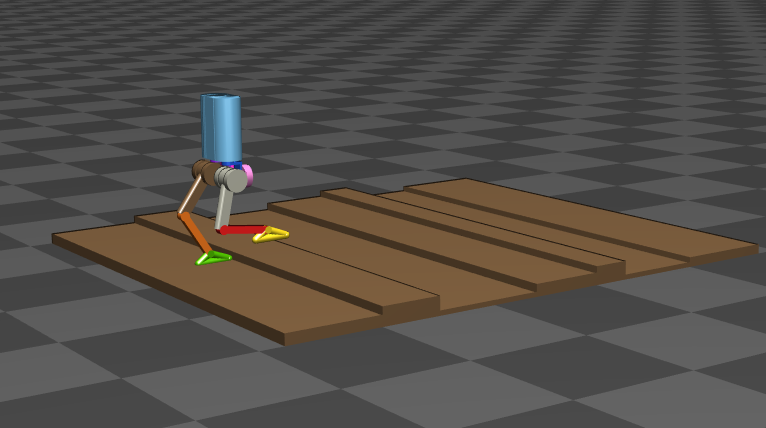}
		\caption{{\bfseries 10-DoF Bipedal Robot.} The bipedal robot walking through rough terrain with velocity $ {{\bm v}_x}_d=1.6\:\unit{m/s}$. Simulation video: \protect\url{https://youtu.be/Z2s4iuYkuvg}. }
		\label{fig:roughTerrainSim}
	\end{figure}
	
MPC has been also utilized in the control of bipedal robots through various approaches. The control framework proposed in  \cite{powell2015model} applies MPC to minimize the values of rapidly exponentially stabilizing control Lyapunov function (RES-CLF) in a Human-Inspired Control (HIC) approach. A revisited ZMP Preview Control scheme presented in \cite{wieber2006trajectory} attempts to solve an optimal control problem by MPC that finds an optimal sequence of jerks of robot center of mass (CoM). 
However, these approaches are either based on position control to track a joint trajectory resulting from optimization; or tend to address the step-by-step planning problem. In this work, we focus on a real-time feedback control approach that can handle a wide range of walking gaits, without relying on offline trajectory optimization.
%\todo{Add 1 paragraph to talk about other approaches for control of bipedal robots such as Hybrid Zero Dynamics. You can check my stepping stones paper for reference.}

Inspired by the successful force-based MPC approach for quadruped robots presented in \cite{di2018dynamic}, in this paper, we propose a new control framework that utilizes MPC to solve for optimal ground reaction forces and moments to achieve dynamic motion on bipedal robots. We investigate different models that can be used for the MPC framework and introduce the formulation that works most effectively on our bipedal robot model. Our proposed approach allows a 10-DOF bipedal robot to perform high-speed and robust locomotion on rough terrain. We implement and validate our controller design in a high-fidelity physical simulation that is constructed in MATLAB and Simulink with the software dependency of Spatial v2.

%\todo{Add 1 paragraph to emphasize the main contribution of our approach here}
The main contributions of the paper are as follows:
\begin{itemize}
    \item We proposed a new framework of force-and-moment-based MPC for 10-DoF bipedal robot locomotion.
    
    \item We investigate different models of rigid body dynamics that can be used for the MPC framework. The most effective model is then used in our proposed approach.
    
    \item The proposed MPC framework allows 3-D dynamic locomotion with accurate velocity tracking.
    \item Our control framework can enable a wide range of dynamic locomotion such as fast walking, hopping, and running using the same set of control parameters. 
    \item Thanks to using the force-and-moment based control inputs, our approach is also robust to rough terrain. We have successfully demonstrated the problem of fast walking with the velocity of $1.6\:\unit{m/s}$ on rough terrain. The rough terrain consists of stairs with a maximum height of $0.075\:\unit{m}$ and a maximum difference of $0.055\:\unit{m}$ between two consecutive stairs.
    % \item The controller design is planned be extended to physical robot testing in the future work.
\end{itemize}

The rest of the paper is organized as follows. Section \ref{sec:robotModel} introduces the model design and physical parameters of the bipedal robot. Simulation methods and control architecture are also provided in this section. Section \ref{sec:dynamicsAndControl} presents the dynamics and controller choices, design, and formulation of the proposed force-and-moment-based MPC controller. Some result highlights are presented in Section \ref{sec:simulationResults} along with an analysis on controller performance in various dynamic motions. 

%% file: Sections/RobotModel.tex
%!TEX root = ../Main.tex

\section{Bipedal Robot Model and Simulation}
\label{sec:robotModel}

\subsection{Robot Model}
\label{subsec:robotModel}

\begin{figure}[!h]
	\hspace{0.2cm}
     \center
     \begin{subfigure}[b]{0.227\textwidth}
         \centering
         \includegraphics[width=\textwidth]{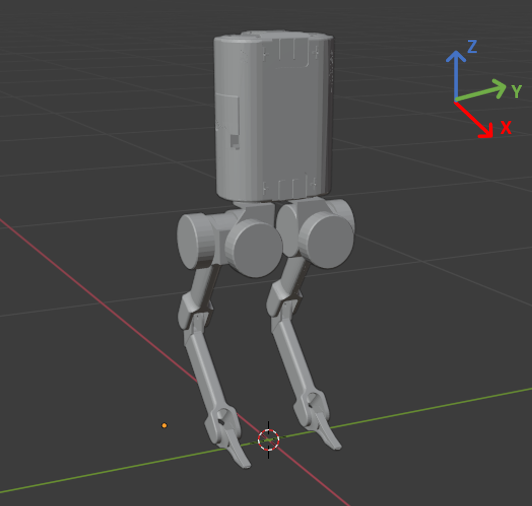}
         \caption{ }
         \label{fig:robotModel}
     \end{subfigure}
     %\hfill
     \begin{subfigure}[b]{0.21\textwidth}
         \centering
         \includegraphics[width=\textwidth]{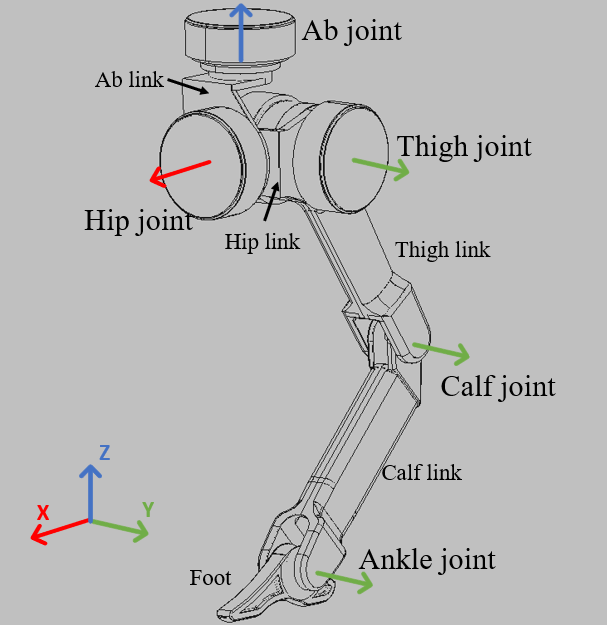}
         \caption{ }
         \label{fig:legConfig}
     \end{subfigure}
        \caption{{\bfseries Bipedal Robot Configuration}  a) Robot CAD Model  b) The link and joint configuration of the bipedal robot left leg.}
        \label{fig:robotConfig}
\end{figure}

In this section, we present the robot model that will be used throughout the paper. A 10 DoF bipedal robot consists of 5 joint actuation each leg (see Fig. \ref{fig:robotModel}). Commonly, a 10 DoF bipedal robot has abduction (ab) and hip joints that allow 3-D rotation and ankle joints that allow double-leg-support standing (e.g., \cite{levineblackbird,gong2019feedback}). 

The design of our bipedal consists of the robot body, ab link, hip link, thigh link, calf link, and foot link (see Fig. \ref{fig:legConfig} for the definition of the leg configuration and Table \ref{tab:PRP} for physical parameters). The robot body is adopted from the Unitree A1 robot, in a vertical configuration. The joint actuator modeled in this robot design is the Unitree A1 modular actuator, which is a lightweight and powerful torque-controlled motor that is suitable for mini legged robots (see Table \ref{tab:motor} for the parameters of the actuator). 

\begin{table}[h]
	\vspace{0.2cm}
	\centering
	\caption{Robot Physical Parameters}
	\label{tab:PRP}
	\begin{tabular}{cccc}
		\hline
		Parameter & Symbol & Value & Units\\
		\hline
		Mass & $m$    & 11.84 & $\unit{kg}$  \\[.5ex]
		Body Inertia  & $I_{xx}$  & 0.0443 & $\unit{kg}\cdot \unit{m}^2$ \\[.5ex]
		& $I_{yy}$ & 0.0535  & $\unit{kg}\cdot \unit{m}^2$ \\[.5ex]
		& $I_{zz}$ & 0.0214  & $\unit{kg}\cdot \unit{m} ^2$ \\[.5ex]
		Body Length & $l_{b}$ & 0.114 & $\unit{m}$ \\[.5ex]
		Body Width & $w_{b}$ & 0.194 & $\unit{m}$ \\[.5ex]
		Body Height & $h_{b}$ & 0.247 & $\unit{m}$ \\[.5ex]
		Thigh and Calf Lengths & $l_{1}, l_{2}$ & 0.2 & $\unit{m}$ \\[.5ex]
		Foot Length & $l_{toe}$ & 0.09 & $\unit{m}$ \\[.5ex]
		& $l_{heel}$ & 0.05 & $\unit{m}$ \\[.5ex]
		\hline 
		\label{tab:robot}
	\end{tabular}
\end{table}	

	\begin{table}[h]
%		\hspace{0.2cm}
		\centering
		\caption{Joint Actuator Parameters}
		\begin{tabular}{cccc}
			\hline
			Parameter & Value & Units\\
			\hline
			Max Torque   &  33.5 & $\unit{Nm}$  \\[.5ex]
			Max Joint Speed    & 21  & $\unit{Rad}/\unit{s}$  \\[.5ex]
			\hline 
			\label{tab:motor}
			\end{tabular}
			\end{table}
\subsection{Simulation}
\label{subsec:simulation}

\begin{figure}[h]
	\hspace{0.2cm}
		\center
		\includegraphics[width=1 \columnwidth]{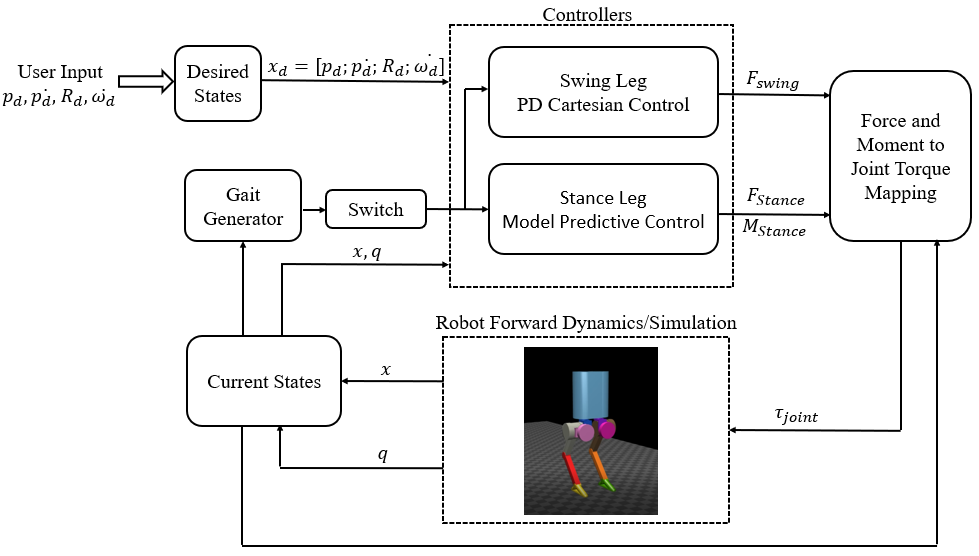}
		\caption{{\bfseries Control Diagram.} The simplified block diagram and control architecture of our proposed approach.}
		\label{fig:controlArchi}
	\end{figure}

The bipedal robot simulation is built primarily in MATLAB Simulink using Spatial v2 package \cite{featherstone2014rigid}. Fig. \ref{fig:controlArchi} shows the diagram for our control architecture and also the representation of our simulation software.
% is an implementation of spatial vector arithmetic and dynamics algorithms that is available in MATLAB scripts. The software employs Roy Featherstone’s book \emph{Rigid Body Dynamics Algorithms} and provides a series of accessible MATLAB functions for robotic dynamics simulations \cite{featherstone2014rigid}. 
% The scope of the simulation construction is to build easy-to-use and reliable MATLAB and Simulink models that can be used as a test bench for many forthcoming controller designs and optimization implementations (see Fig. \ref{fig:controlArchi}: simulation block diagram).

The simulation software requires the user to input desired states at the beginning of the simulation. The desired states form a column vector that consist desired body center of mass (CoM) position $\bm p_d$, desired body CoM velocity $\bm {\dot p}_d$, desired rotation matrix $\bm R_d$ (resized to 9×1 vector), and desired angular velocity $\bm {\omega}_d$ of robot body. 

% The simulation model shown in Fig. \ref{fig:controlArchi} provides a platform for controller implementation with the advantages of fast simulation time, simplicity in modification and debugging, and reliability.

%% file: Sections/DynamicsAndControl.tex
\section{Dynamics and Control}
\label{sec:dynamicsAndControl}
\subsection{Simplified Dynamics}
\label{subsec:simplified_dynamics}
In this Section, we investigate different dynamic models that can be used in our MPC control framework. 
While the whole-body dynamics of the bipedal robot are highly non-linear, we are interested in using simplified rigid-body dynamics to guarantee that our MPC controller can be solved effectively in real-time. 
In addition, in order to enable the capability of absorbing frequent and hard impacts from dynamic locomotion, the design of bipedal robots also requires lightweight limbs and connections. This has allowed the weight and rotation inertia of each link part to be very small compared to the body weight and rotation inertia. Hence, the effect of leg links in the robot dynamics may be neglected, forming simplified rigid-body dynamics \cite{nguyen2019optimized,bledt2018cheetah}. This is also a common assumption in many legged robots’ controller designs \cite{focchi2017high,stephens2010push}.

\begin{figure}[!h]
\vspace{0.2cm}
     \centering
     \begin{subfigure}[b]{0.15\textwidth}
         \centering
         \includegraphics[width=\textwidth]{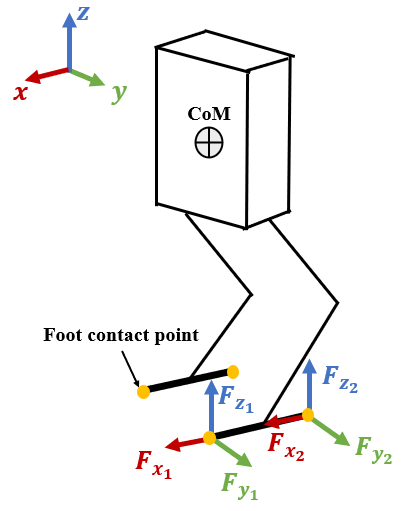}
         \caption{Model 1}
         \label{fig:model1}
     \end{subfigure}
     \hfill
     \begin{subfigure}[b]{0.13\textwidth}
         \centering
         \includegraphics[width=\textwidth]{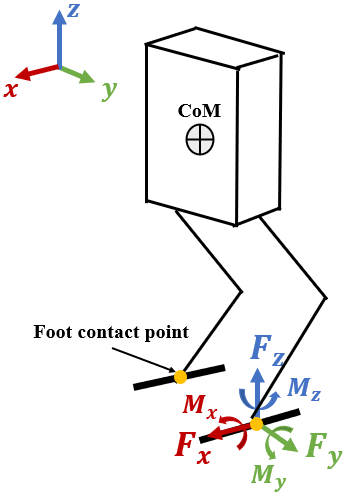}
         \caption{Model 2}
         \label{fig:model2}
     \end{subfigure}
     \hfill
     \begin{subfigure}[b]{0.145\textwidth}
         \centering
         \includegraphics[width=\textwidth]{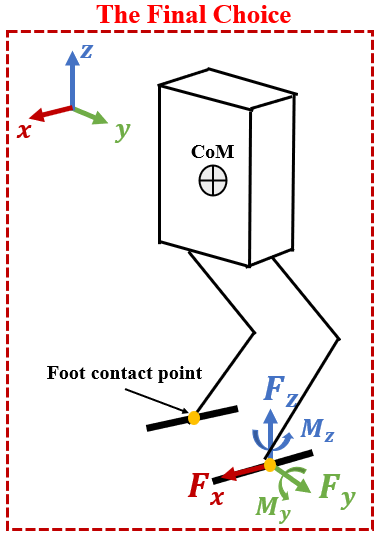}
         \caption{Model 3}
         \label{fig:model3}
     \end{subfigure}
        \caption{{\bfseries Three Simplified Dynamics Models.} Yellow dots on the figures represent the contact point locations in simplified dynamics a) 2 contact points at the toe and heel of each robot foot, 3-D forces applied  b) 1 contact point at the middle of each foot, 3-D force and 3-D moments applied  c) 1 contact point at the middle of each foot, 3-D force and 2-D moments applied. Model 3 is the final choice to be used in our proposed approach.}
        \label{fig:simplifiedDynamics}
\end{figure}

There are three simplified dynamics models that we have considered and tested, shown in Fig. \ref{fig:simplifiedDynamics}. The main difference between these three options is the number of contact points on each foot, contact location, and contact force and moment formation at each contact point. Model 1 resembles the simplified dynamics choice for quadruped robots mentioned in \cite{bledt2018cheetah}, with 4 contact points exerting 3-D contact forces. However, with this dynamics model applied to our 10-DoF bipedal robot, under rotation motions testings in simulation, the robot is unable to perform pitch motion properly. Model 2 is improved and further simplifies the contact points. However, with this simplified dynamics model, in simulation validation process, the robot is unable to perform roll motion correctly. Hence, we proposed model 3 that excludes external moment around $x$-axis, and contains only 3-D contact forces and 2-D moments around $y$ and $z$-axis. This model has allowed the robot to perform all 3-D rotations effectively. Hence, model 3 is chosen to be the final simplified dynamics design in this framework. More details about the validation of this decision process is shown in Section \ref{sec:simulationResults}.
The detailed derivation of the model 3 dynamics is presented as follows.

The bipedal model in this paper has two legs that both consists of 5 DoF. Commonly, the external contact forces applied to the robot are only limited to 3-D forces in many legged-robot dynamics (e.g., \cite{levineblackbird,nguyen2019optimized}). However, thanks to the additional hip and ankle joint actuation, the external moments can also be included in the robot dynamics, forming a linear relationship between robot body’s acceleration $ {\bm {\ddot p}_c}$, rate of change in angular momentum $\bm {\dot H}$ about CoM \cite{stephens2010push}, and contact force and moments %$ \bm F=[ \bm F_1,\:\bm F_2]^T$ , where $\bm F_i= [\bm F_{ix},\:\bm F_{iy},\:\bm F_{iz},\:\bm M_{ix},\:\bm M_{iy},\:\bm M_{iz} ]^T$, $i=1,2$ , as follows:
\update{$ \bm u=[\bm F_1,\:\bm F_2,\:\bm M_1,\:\bm M_2]^T,  \bm F_i = [\bm F_{ix},\:\bm F_{iy},\:\bm F_{iz}]^T, \bm M_i = [\bm M_{iy},\:\bm M_{iz}]^T, i = 1, 2, $ shown as follows: 
\setlength{\belowdisplayskip}{5pt} \setlength{\belowdisplayshortskip}{5pt}
\setlength{\abovedisplayskip}{5pt} \setlength{\abovedisplayshortskip}{5pt}
\begin{align}
\label{eq:simplifiedDynamics}
\left[\begin{array}{ccc} \bm {{D}_1}   \\ 
 \hspace{0cm} \bm {{D}_2} \end{array}  \right] \bm u= \left[\begin{array}{c} m (\ddot{\pos}_{c} +\bm{g}) \\ \bm {\dot H} \end{array} \right],
\end{align}
where
\begin{align}
\label{eq:D1}
 \bm {D}_1  = \left[\begin{array}{cccc} \mathbf {I}_{3\times3}  & \mathbf {I}_{3\times3} & 
  \mathbf {0}_{3\times2} &  \mathbf {0}_{3\times2} \end{array} \right],
\end{align}
\begin{align}
\label{eq:D2}
 \bm {D}_2  = \left[\begin{array}{cccc} \bm {(p_1-p_c)}\times  & \bm {(p_2-p_c)}\times & 
  \mathbf L &  \mathbf L \end{array} \right],
\end{align}
\begin{align}
\label{eq:L}
 \mathbf L  = \left[\begin{array}{ccc}  \mathbf 0 & \mathbf 0 \\  \mathbf 1 &\mathbf 0 \\  \mathbf 0 &\mathbf 1
 \end{array} \right].
\end{align} }

The term ${(\bm p_i- \bm p_c)}$ denotes the distance vector from the robot body CoM location to the foot $i$ position in the world coordinate; and $\bm {(p_i-p_c)}\times $ represents the skew-symmetric matrix representing the cross product of ${ {(\bm p_i-\bm p_c)} \times \bm F_i }$. Here, $\bm {\dot H}$ can be approximated as $\bm{ \dot H=I_G \dot {{\omega}}}$  (as discussed in \cite{nguyen2019optimized}), where $\bm {I_G}$ stands for the centroid rotation inertia of robot body in the world frame and  $\bm {\dot {\omega}}$ represents the angular velocity of robot body in the world frame \cite{bledt2018cheetah,stephens2010push}.

%\todo{We should have a figure to illustrate the simplified rigid body model including 1 rigid body, 2 contact point on the ground, force vectors, My, Mz,...}
%\todo{We should also consider make comparisons between different models: the current one (1 contact point each leg: 3 forces + 2 moments for each point), 2 contact point each leg (3 forces + 3 moments for each points), 2 contact point each leg (3 forces + 0 moments for each points)... like you used to explore. This will make the paper stronger.}
\update{Equation (\ref{eq:simplifiedDynamics}) to (\ref{eq:L}) describes the simplified dynamics model 3. This model only used 5 force and moment inputs $\bm U$ which are directly mapped to 5 joint torques in each leg.}

We use rotation matrix $\bm R$ as a state variable to represent the orientation of the robot body, which can be also directly converted to Euler Angles. 
%Reference \cite{di2018dynamic} presents the 
We linearize the rotation matrix by approximating the angular velocity in terms of Euler angle $ {\Theta = [\phi,\:\theta,\:\psi]}^T$, where $\phi$ is roll angle,  $ \theta$ is pitch angle, and $ \psi$ is yaw angle. With the assumption of small roll and pitch angles  \cite{di2018dynamic}, the relation of the rate of change of $\Theta$ and angular velocity $\bm \omega$ in the world coordinate can be approximated as: 
%\begin{align}
%\label{eq:omega}
%%\bm \omega  = \left[\begin{array}{ccc} {\cos{\theta} \cos{\psi}}& -{\sin{\psi}} & 0 \\{\cos{\theta} \sin{\psi}} & {\cos{\psi}} & 0 \\ -{\sin{\theta}} & 0 & 1\end{array} 
%\right ] 
%\begin{bmatrix} {\dot{\phi}} \\ {\dot{\theta}} \\ {\dot{\psi}}\end{bmatrix}
%\end{align} 

%\begin{align}
%\label{eq:omega}
%\bm \omega  = \left[\begin{array}{ccc} %\dot{\phi}\sin{\theta}\sin{\psi}+\dot{\theta}\cos{\psi} \\ 
%\dot{\phi}\sin{\theta}\cos{\psi}-\dot{\theta}\sin{\psi} \\ 
%\dot{\phi}\cos{\theta}+\dot{\psi}
%\end{array} 
%\right ] 
%\end{align} 

%With small pitch and roll angles, equation (\ref{eq:omega}) can be approximated and rewritten as
\begin{align}
\label{eq:omega2}
\begin{bmatrix} {\dot{\phi}} \\ {\dot{\theta}} \\ {\dot{\psi}}\end{bmatrix} \approx \left[\begin{array}{ccc} { \cos{\psi}}& {\sin{\psi}} & 0 \\{ -\sin{\psi}} & {\cos{\psi}} & 0 \\ 0 & 0 & 1\end{array} 
\right ] 
\begin{array}{cc}
     {\bm \omega}
\end{array}.
\end{align} 
% \todo{You should double-check eq 5,6 because they may not be correct in [9].}

Hence the kinematic constraint of the Euler angles is obtained as follow:
\begin{align}
\label{eq:omega3}
\begin{bmatrix} {\dot{\phi}} \\ {\dot{\theta}} \\ {\dot{\psi}}\end{bmatrix} \approx \bm R_z(\psi) \bm \omega.
\end{align} 

Combing the approximated orientation dynamics and the translation dynamics, the simplified dynamics of the robot can be written as: 
\update{
\begin{align}
\label{eq:stateSpace}
\frac{d}{dt}
\begin{bmatrix} 
{\bm \Theta} \\
{\bm p}_c \\ 
{\bm \omega} \\ 
\dot {{\bm p}}_c
\end{bmatrix} = 
\bm A
\begin{bmatrix} 
{\bm \Theta} \\
{\bm p}_c \\ 
{\bm \omega} \\ 
\dot {{\bm p}}_c
\end{bmatrix} 
+ \bm B \bm u +
\begin{bmatrix} 
\mathbf 0 \\ \mathbf 0 \\\mathbf  0 \\\mathbf  g
\end{bmatrix},
\end{align} 
where
\begin{align}
\label{eq:A}
\bm A = 
\begin{bmatrix} 
\mathbf {0}_{3\times3} & \mathbf {0}_{3\times3} & \bm R_z(\psi) & \mathbf {0}_{3\times3}  \\
\mathbf {0}_{3\times3} & \mathbf {0}_{3\times3} & \mathbf {0}_{3\times3} & \mathbf  {I}_{3\times3}  \\ 
\mathbf {0}_{3\times3} & \mathbf {0}_{3\times3} & \mathbf {0}_{3\times3} & \mathbf {0}_{3\times3} \\ 
\mathbf {0}_{3\times3} & \mathbf {0}_{3\times3} & \mathbf {0}_{3\times3} & \mathbf {0}_{3\times3}, 
\end{bmatrix},
\end{align} 
\begin{align}
\setlength\arraycolsep{2.5pt}
\label{eq:B}
\bm B = 
\begin{bmatrix} 
\mathbf {0}_{3\times3} & \mathbf {0}_{3\times3} & \mathbf {0}_{3\times2} & \mathbf {0}_{3\times2}  \\
\mathbf {0}_{3\times3} & \mathbf {0}_{3\times3} & \mathbf {0}_{3\times2} & \mathbf {0}_{3\times2}  \\ 
\bm {{\Tilde{I}^{-1}_G} (p_1-p_c)}\times & \bm {{\Tilde{I}^{-1}_G} (p_2-p_c)}\times & \bm{{\Tilde{I}^{-1}_G} \mathbf L} & \bm{{\Tilde{I}^{-1}_G} \mathbf L} \\ 
\mathbf { {I}_{3\times3}}/m & \mathbf {{I}_{3\times3}}/m & \mathbf {0}_{3\times2} & \mathbf {0}_{3\times2} 
\end{bmatrix}
\end{align} }

\update{Here, $\bm{\Tilde{I}^{-1}_G}$ is approximated by rotation inertia of the robot body in its body frame $\bm I_b$ and $\bm R_z(\psi)$ from \eqref{eq:omega3}:
\begin{equation}
\label{eq:rotationI}
\bm{\Tilde{I}_G} = \bm R_z(\psi) \bm I_b {\bm R_z(\psi)}^T.
\end{equation}
}
By assigning gravity as additional state variable, now state ${{\bm x}} = [{{\bm \Theta}},\: {\bm p}_c,\:{{\bm \omega}},\:\dot { {\bm p}}_c,\:\bm g]^T$ will allow the dynamics in (\ref{eq:stateSpace}) to be rewritten into a linear state-space form with continuous time matrices $\bm {\hat{A_c}}$ and $\bm {\hat{B_c}}$: 
\begin{equation}
\label{eq:linearSS}
\dot{{\bm { x}}}(t) = \bm {\hat{A_c}}(\psi) {{\bm {x}}}(t) + \bm {\hat{B_c}}([p_1-p_c],[p_2-p_c],\psi) \bm u(t).
\end{equation}

The linearized dynamics in (\ref{eq:linearSS}) is now suitable for the convex MPC formulation presented in \cite{di2018dynamic}.

\subsection{MPC Formulation}
\label{subsec:MPC}
Having discussed the dynamics model, we now present details about the formulation of our MPC controller.

The linearized dynamics in (\ref{eq:linearSS}) can be represented in a discrete-time form at each time step $i$
\begin{align}
\label{eq:discreteDynamics}
{\bm {x}}[i+1] = \bm {\hat{A}}[i] \bm x[i] + \bm {\hat{B}}[i]\bm u[i],
\end{align}
where discrete time matrix $\bm {\hat{A}}$ is a constant matrix computed from $\bm {\hat{A_c}}(\psi)$ using a average yaw value during entire reference trajectory; and $\bm {\hat{B}}$ matrix is computed from $ \bm{\hat{B_c}}([p_1-p_c],[p_2-p_c],\psi)$, using the desired values of average yaw and foot location. The only exception is that at the first time step,  $\bm {\hat{B}}[1]$ is computed from current states of the robot instead of reference trajectory.

An MPC problem with a finite horizon length $k$ can be written in the following standard form:
\begin{align}
\label{eq:MPCform}
%\nonumber
\underset{\bm{x,u}}{\operatorname{min}}   \:\:  & \sum_{i = 0}^{k-1}(\bm x_{i+1}-  {\bm x_{i+1}}_{ref})^T\bm Q_i(\bm x_{i+1}- {\bm x_{i+1}}_{ref}) + \| \bm{u}_i \|\bm{R}_i
\end{align}
\begin{align}
\label{eq:dynamicCons}
\:\:\mbox{s.t. }& \quad  {\bm {x}}[i+1] = \bm {\hat{A}}[i]\bm x[i] + \bm {\hat{B}}[i]\bm u[i], \quad i = 0 \dots k-1
\end{align}
\begin{align}
\label{eq:MPCineqCons}
\quad  \bm {c^-}_i \leq \bm C_i\bm u_i \leq \bm {c^+}_i, \quad i = 0 \dots k-1
\end{align}
\begin{align}
\label{eq:MPCeqCons}
& \quad \bm D_i \bm u_i = 0 , \quad i = 0 \dots k-1
\end{align}

%\todo{We should use the format $x^TQx$ instead of $||x||Q$ in the cost function.}

In (\ref{eq:MPCform}), $\bm x_i$ and $\bm u_i$ are system states and control inputs at time step $i$. \update{Note that the MPC prediction is computed based on the measured states of current step (i.e. $i = 0$).} $\bm Q_i$ and $\bm R_i$ are matrices defining the weights of \update{each state and control input variable.} $\bm {\hat{A}}$ and $\bm {\hat{B}}$ in (\ref{eq:dynamicCons}) are the discrete-time system dynamic constraints from (\ref{eq:discreteDynamics}). $\bm {c^-}_i$,$\bm {c^+}_i$, and $\bm C_i$ in (\ref{eq:MPCineqCons}) represents the inequality constraints of the MPC problem. $\bm D_i$ in (\ref{eq:MPCeqCons}) represents the equality constraints. In this problem, the equality constraint governs the optimal control input from MPC controller is a zero vector for swing foot. 

The MPC controller solves the optimal ground contact force and moment with respect to dynamic constraints \eqref{eq:dynamicCons} and the following inequality constraints:
\begin{align}
\label{eq:frictionCons}
\nonumber 
-\mu \bm {F}_{iz} \leq \bm F_{ix} \leq \mu \bm {F}_{iz} \\
-\mu \bm {F}_{iz} \leq \bm F_{iy} \leq \mu \bm {F}_{iz} \\
% \end{align}
% \begin{align}
\label{eq:forceCons}
0< \bm {F}_{min} \leq \bm  F_{iz} \leq \bm {F}_{max} \\
% \end{align}
% \begin{align}
\label{eq:torqueCons}
\bm |{\tau}_{i}| \leq \bm {\tau}_{max}. 
\end{align}
Here, \eqref{eq:frictionCons} governs the contact forces in $x$ and $y$ direction are within the friction pyramid, with $\mu$ being the friction coefficient. The contact forces in $z$-direction should also fall within the upper and lower bounds of force (\ref{eq:forceCons}), where the lower bound is positive to maintain contact with the ground. It is also important to restrict the joint torques to be within the saturation of the physical motor (\ref{eq:torqueCons}).

\subsection{QP Formulation}
% It is stated in Section \ref{sec:robotModel} of this paper that the scope of developing the robot simulation in MATLAB and Simulink is to have a fast and reliable simulation software to test controller designs. MPC problem can be heavy to solve so it is important to reduce the computation and problem size of MPC by formulating the problem into a quadratic program (QP).
With the linear dynamics in Section \ref{subsec:simplified_dynamics} and the MPC formulation in Section \ref{subsec:MPC}, our controller can be formulated as a quadratic program (QP) that can be solved effectively in real-time. 
% With given equality and inequality constraints, we can form a QP problem based on the dynamics from the condensed formulation in (\ref{eq:QPdynamics}). The Optimization Toolbox in MATLAB provides powerful and fast QP solver which is suitable for this problem. 

Firstly, the dynamic constraints \eqref{eq:linearSS} for the entire MPC prediction horizon can be written as:
\begin{align}
\label{eq:QPdynamics}
\bm X = \bm{A}_{qp} \bm x_0 + \bm{B}_{qp} \bm U,
\end{align}
where $\bm X$ is a column vector containing system states for the next $k$ horizons, $\bm x[i+1],\bm x[i+2] \dots {\bm x}[i+k]$ and $\bm U$ is a column vector containing optimal control inputs of current state $\bm u[i]$ and next $k-1$ horizons, $\bm u[i+1], \bm u[i+2] \dots {\bm u}[i+k-1]$ at time step $i$. 
The MPC controller now can be written as the following QP form:     
\begin{align}
\label{eq:QPform}
%\nonumber
\underset{\bm{U}}{\operatorname{min}}   \:\:  & \frac{1}{2}\bm U^T\bm h \bm U + \bm U^T\bm f \\
% \end{align}
% \begin{align}
\label{eq:QPineqCons}
\:\:\mbox{s.t. }& \quad  \bm C\bm U \leq \bm d \\
% \end{align}
% \begin{align}
\label{eq:QPeqCons}
& \bm A_{eq}\bm U = \bm b_{eq}
\end{align}
where $\bm C$ and $\bm d$ are inequality constraint matrices, $\bm A_{eq}$ and $\bm b_{eq}$ are equality constraint matrices, and 
\begin{align}
\label{eq:h}
\bm h = 2( {\bm B_{qp}}^T\mathbf M {\bm B_{qp}}+\mathbf K), \\
% \end{align}
% \begin{align}
\label{eq:f}
\bm f = 2 {\bm B_{qp}}^T\mathbf M ({\bm A_{qp}}\bm x_0-\bm y).
\end{align}
Diagonal matrices $\mathbf K$ and $\mathbf M$ are the weights for the rate of change of state variables and force/moment magnitude.

\update{The resulting controller input of each leg from QP problem $\bm u_i=[\bm F_i,\: \bm M_i]^T$ is mapped to its joint torques by
\begin{align}
\label{eq:forceTorquemap}
\bm {\tau}_i =  \bm J_i^T \bm u_{i},
\end{align} 
where $\bm J_i$ is the Jacobian matrix of $i$th leg 
\begin{align}
\label{eq:Jacobian}
\bm {J}_i^T = \begin{bmatrix} 
\bm {J}_v^T & \bm {J}_\omega^T \bm L
\end{bmatrix},
\end{align} 
}
with $\bm {J}_v $ and $\bm {J}_{\omega} $ being the linear velocity and angular velocity components of $\bm {J}_i $.

\subsection{Swing Leg Control}
As discussed earlier in this section, due to equality constraints, the robot leg that is under the swing phase does not exert ground contact forces and therefore is not under the control of force-and-moment-based MPC. In order to control the leg and foot position in each gait cycle, the desired foot trajectory is under control in the Cartesian space with a PD position controller. \update{The gait sequence is purely based on timing and the gait cycle length is currently set at $0.3 \unit{s}$.}
We obtain the current foot location using forward kinematics. Foot velocity is computed by:
\begin{align}
\label{eq:footVel}
{\dot {\bm p}}_{{foot}_i} =  \bm J_i^T \dot{\bm q_i},
\end{align}
where $\dot{\bm q_i}$ is the joint velocity state-feedback of each leg at time step $i$.

The desired foot location $\bm p_{{foot}_d}$ in the world frame is determined by the foot placement policy employed in \cite{di2018dynamic}:
\begin{align}
\label{eq:footPlacement}
\bm p_{{foot}_d} =  \bm p_{hip} + \dot{\bm p}_c \Delta t/2,
\end{align}
where $\bm p_{hip}$ is the hip joint location in the world frame and $\Delta t$ is the time that stance foot spends on the ground during one gait cycle. 

The swing leg force can be computed by treating the foot attached to a virtual spring-damper system \cite{chen2020virtual}. The foot weight is reasonable to be neglected since it is very small compared to the robot body \cite{nguyen2019optimized}. Following the PD control law, the foot force can be written as: 
\begin{align}
\label{eq:PDswing}
\bm F_{swing_i}=\bm K_P(\bm p_{{foot}_d}-\bm p_{{foot}_i})+\bm K_D(\dot{\bm p}_{{foot}_d}-\dot{\bm p}_{{foot}_i})
\end{align}
where $\bm K_P$ and $\bm K_D$ are PD control gains, or spring stiffness and damping coefficient of the virtual spring-damper system. %$\bm p_{{foot}_d}$ and $\dot{\bm p}_{{foot}_d}$ are desired foot placement and velocity, respectively.

Similar to (\ref{eq:forceTorquemap}), the joint torque can be computed by:
\begin{align}
\label{eq:forceTorqueMapSwing}
\bm {\tau}_{swing_i} =  \bm J_v^T \bm F_{swing_i}.
\end{align}

With Cartesian PD control, the swing leg can move and be controlled to follow desired foot placement trajectory. The gait generator decides either the robot leg is in the stance phase or swing phase in a fixed gait cycle and assigns the appropriate controller to the corresponding leg. Now the robot has both swing and stance leg control, it is ready to test the MPC in simulation.

%% file: Sections/SImulationResults.tex
\section{Simulation Results}
\label{sec:simulationResults}

In this section, we present numerical validation of our proposed approach for different dynamic locomotion. 
% The simulation video for this paper is given in Fig. \ref{fig:roughTerrainSim}. 
The reader is encouraged to watch the supplemental video\footnote{\url{https://youtu.be/Z2s4iuYkuvg}} 
for the visualization of our results.
For our simulation, the bipedal robot model and ground contact model are set up in MATLAB with Spatial v2 software. The MPC sampling frequency is set to $0.03\:\unit{s}$ while the simulation is run at $1\: \unit{kHz}$. One gait cycle that contains 10 horizons is predicted at each time step in MPC, in which each gait cycle is fixed at $0.30\:\unit{s}$. \update{This prediction length has been also used in \cite{di2018dynamic}.}

\update{The weighting factors $\bm Q$ in \eqref{eq:MPCform} are tuned to balance the performance between different control actions. In our simulation, we use $\bm Q_x = \bm Q_y = 50$, $\bm Q_z = 100$, $\bm Q_{\phi} = \bm Q_{\theta} = 100$, and $\bm Q_{\psi} = 20$. The rest weighting factors in $\bm Q$ remains at 1. }

\begin{figure}%[!h]
\vspace{0.5cm}
     \centering
     \begin{subfigure}[b]{0.13\textwidth}
         \centering
         \includegraphics[width=\textwidth]{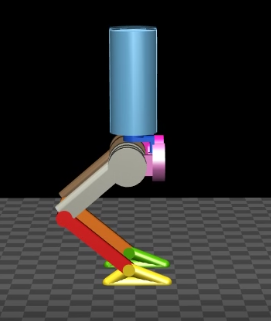}
         \caption{Snapshot of Model 1 in Double-leg Stance}
         \label{fig:pitchModel1}
     \end{subfigure}
     %\hfill
     \quad \quad
     \begin{subfigure}[b]{0.13\textwidth}
         \centering
         \includegraphics[width=\textwidth]{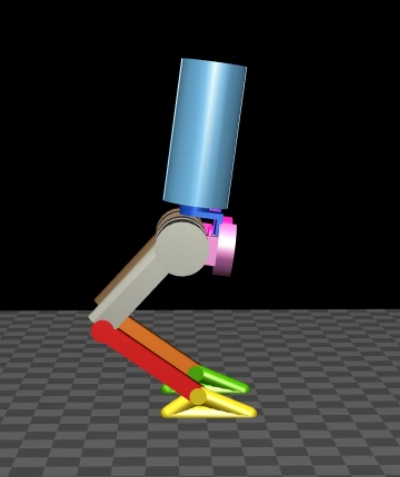}
         \caption{Snapshot of Model 3 in Double-leg Stance}
         \label{fig:pitchModel3}
     \end{subfigure}
     \hfill
     \begin{subfigure}[b]{0.5\textwidth}
         \centering
         \includegraphics[width=0.9\textwidth]{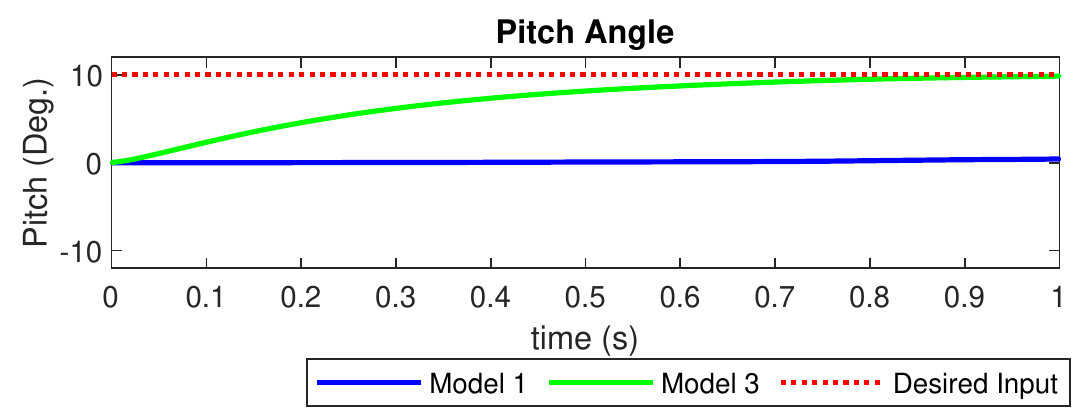}
         \caption{Pitch Motion Comparison}
         \label{fig:pitchPlot}
     \end{subfigure}
        \caption{{\bfseries Comparison of Model 1 and Model 3 in Pitch Motion Simulation}  a) Snapshot at the end of simulation with model 1  b) Snapshot at the end of simulation with model 3   c) Pitch motion response comparison with a $10^{\circ}$ desired pitch input.}
        \label{fig:pitchComparison}
\end{figure}

\begin{figure}%[!h]
	\hspace{0.2cm}
     \center
     \begin{subfigure}[b]{0.13\textwidth}
         \centering
         \includegraphics[width=\textwidth]{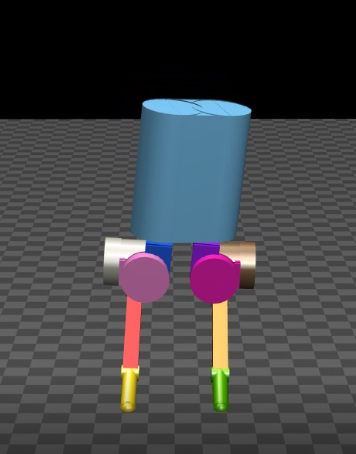}
         \caption{Snapshot of Model 2 in Double-leg Stance}
         \label{fig:rollModel2}
     \end{subfigure}
     %\hfill
     \quad \quad
     \begin{subfigure}[b]{0.14\textwidth}
         \centering
         \includegraphics[width=\textwidth]{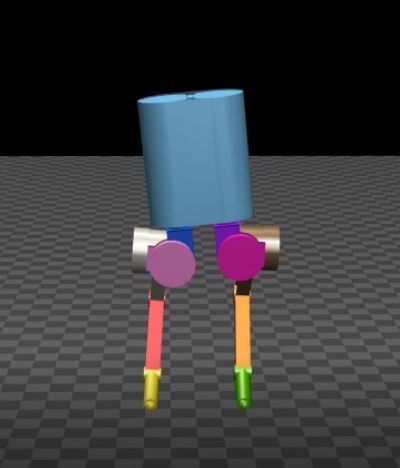}
         \caption{Snapshot of Model 3 in Double-leg Stance}
         \label{fig:rollModel3}
     \end{subfigure}
     %\hfill
     \begin{subfigure}[b]{0.5\textwidth}
         \centering
         \includegraphics[width=0.9\textwidth]{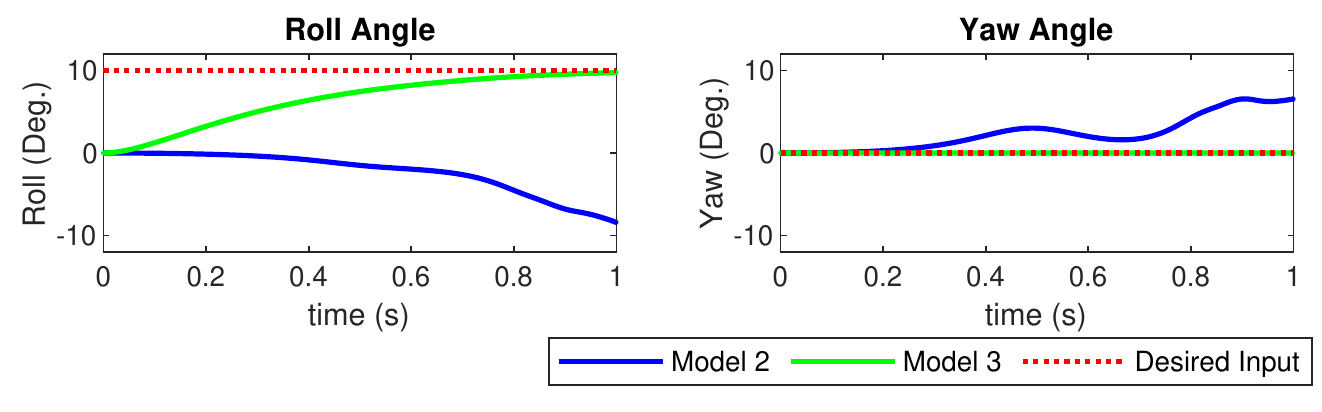}
         \caption{Roll Motion Comparison}
         \label{fig:rollPlot}
     \end{subfigure}
        \caption{{\bfseries Comparison of Model 2 and Model 3 in Roll Motion Simulation}  a) Snapshot at the end of simulation with model 2  b) Snapshot at the end of simulation with model 3   c) Roll motion response comparison with a $10^{\circ}$ desired roll input.}
        \label{fig:rollComparison}
\end{figure}

\subsection{Validation of Simplified Dynamics}
First, we present the simulation results of simple rotation motions during standing with both legs on the ground to validate the claim in Section \ref{sec:dynamicsAndControl} that for the simplified dynamics used for control design, model 3 is a superior choice over model 1 and model 2. 

As mentioned in Section \ref{sec:dynamicsAndControl}, the simplified dynamics model 1 is unable to perform pitch motion. It is shown in Fig. \ref{fig:pitchComparison}, a pitch motion comparison between using simplified dynamics model 1 and model 3. The latter one is what we ultimately chose to use in MPC formulation. It is observed that the simulation result with model 1 does not respond to desired pitch input, whereas model 3 can perform pitch motion. 

We then further simplified model 1 and added 3-D moment inputs to each contact point to form simplified dynamics model 2. However, in the roll motion test, the response with model 2 is incorrect to desired roll input and it also shows a deviation in yaw angle as shown in Fig. \ref{fig:rollComparison}. With model 3, the robot simulation succeed in the roll motion test. 
Therefore, we decide to use model 3 for our proposed approach. Following are simulation results for walking and hopping motion using MPC control for model 3.

\subsection{Velocity Tracking}

\begin{figure}[h]
	\hspace{0.2cm}
	\center
	\includegraphics[width=0.75 \columnwidth]{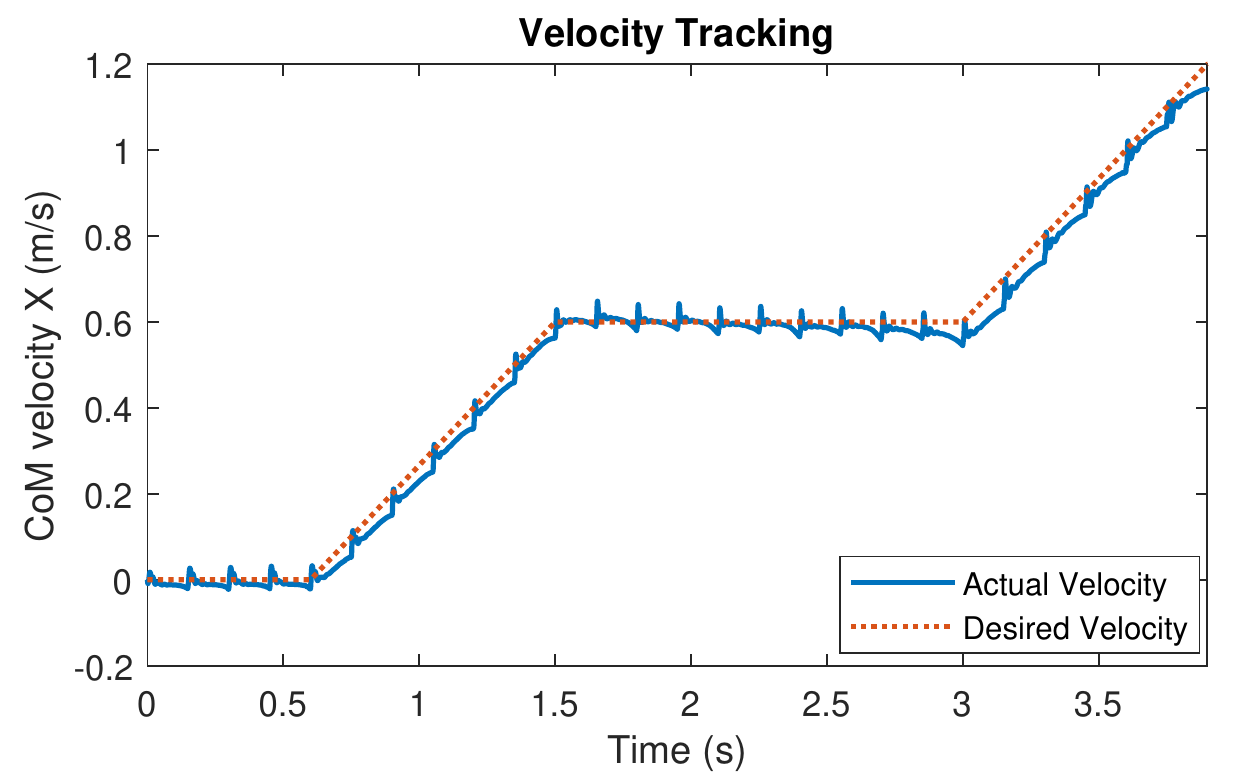}
	\caption{{\bfseries Velocity Tracking.} Comparison of desired velocity input and actual velocity response in x-direction. 
% 	The desired velocity curve keeps constant from $t=0\:\unit{s}$ to $t=0.6\:\unit{s}$ and from $t=1.5\:\unit{s}$ to $t=3\:\unit{s}$ at $\bm v_{x_d}=0\:\unit{m/s}$ and $\bm v_{x_d}=0.6\:\unit{m/s}$, respectively. From $t=0.6\:\unit{s}$ to $t=1.5\:\unit{s}$ and from $t=3\:\unit{s}$ to $t=3.9\:\unit{s}$, $\bm v_{x_d}$ increases linearly.
	}
	\label{fig:velTracking}
\end{figure}

In this simulation, we test the MPC performance in forward walking motion(positive $x$-direction) with time-varying desired speed and the desired CoM height of $0.5\:\unit{m}$. 
%The joint torques during this simulation are presented in Fig. \ref{fig:velSimTorque}. Note that during the simulation, all joint torque data are within the maximum torque threshold of our motor choice. There are no extreme torque values found during this entire simulation. 
The velocity tracking plot is shown in Fig. \ref{fig:velTracking}, the actual response curve with MPC shows a good tracking performance. The velocity response has a maximum deviation of $0.076\: \unit{m/s} $ compared to the desired input. Besides walking forward, we also have successful simulation results and demonstrations in walking sideways and diagonally. This result validates the effectiveness of our proposed control framework in realizing 3D dynamic locomotion for bipedal robots. 
% The simulation results can be found in the video (URL is under Fig. \ref{fig:roughTerrainSim}) associated with this paper. 
%\begin{figure}[h]
%	\center
%	\includegraphics[width=1 \columnwidth]{Figures/MPC_tau.png}
%	\caption{{\bfseries Plots of Joint Torques with MPC.} Joint torques of stance leg under the control of MPC in time-varying velocity simulation. }
%	\label{fig:velSimTorque}
%\end{figure}
	
\subsection{High-velocity Walking in Rough Terrain}
We also validated the controller performance in rough terrain locomotion at high speed. Specifically, the robot is commanded to walk through a $2.4$-meter-long rough terrain formed by stairs with various heights and lengths. The stair heights range from $0.020\:\unit{m}$ to $0.075\:\unit{m}$ with a maximum height difference of $0.055\:\unit{m}$ between two consecutive stairs. To validate the feasibility and potential of MPC locomotion through rough terrain, the robot is commanded to follow a high desired velocity $\bm v_{x_d}=1.6 \:\unit{m/s}$. A snapshot of this simulation is provided in Fig. \ref{fig:roughTerrainSim}. 
%\begin{figure}[h]
%	\center
%	\includegraphics[width=0.85 \columnwidth]{Figures/Rough_terrain_spatial.png}
%	\caption{{\bfseries Snapshot from Rough Terrain Simulation.} Terrain model with stairs at various heights, animated with Spatial v2.  }
%	\label{fig:roughTerrain}
%\end{figure}

\begin{figure}[h]
	\hspace{0.2cm}
	\center
	\includegraphics[width=0.95 \columnwidth]{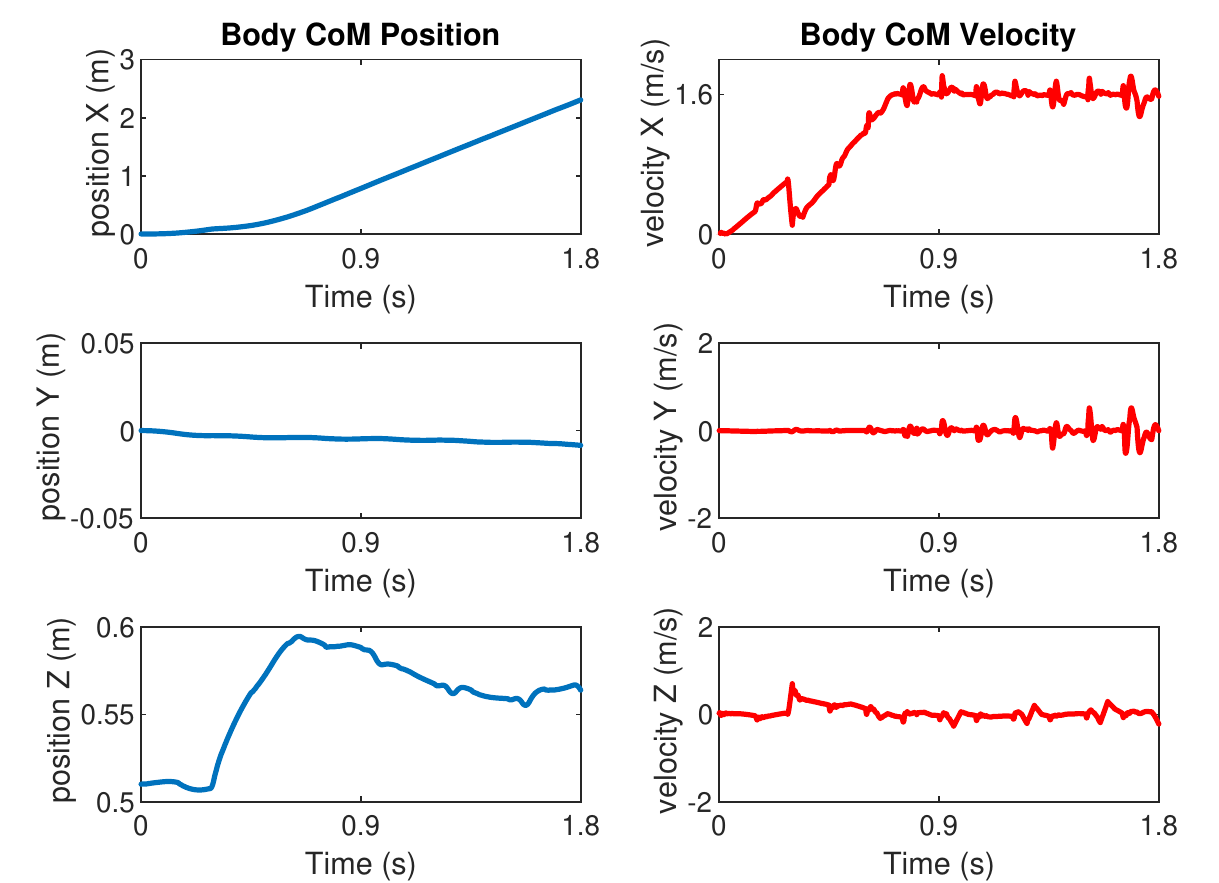}
	\caption{{\bfseries Plots of Body CoM Position and Velocity in Rough Terrain Simulation.}  }
	\label{fig:CoMposVel}
\end{figure}

\begin{figure}[h]
	\center
	\includegraphics[width=0.88 \columnwidth]{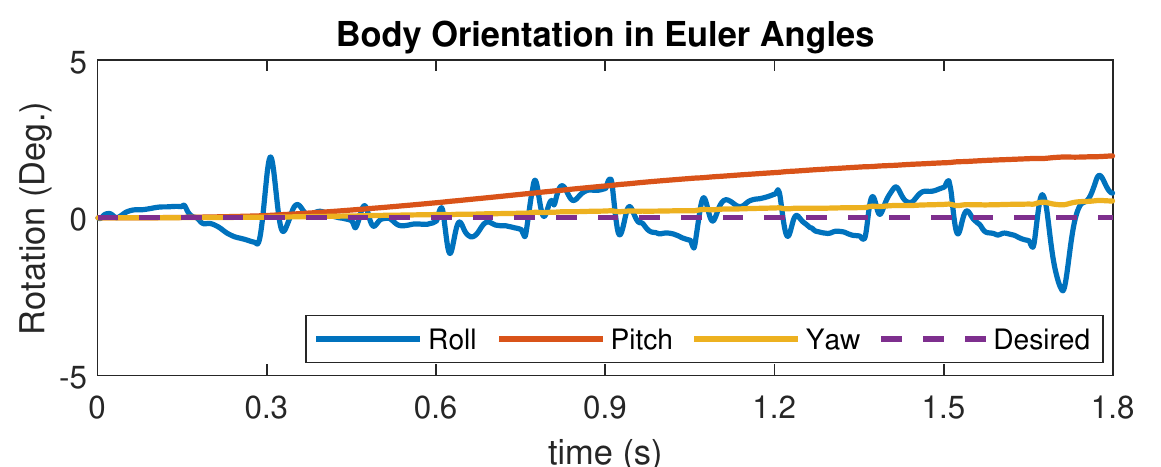}
	\caption{{\bfseries Plots of Robot Orientation in Rough Terrain Simulation. }  }
	\label{fig:CoMeulAng}
\end{figure}

\begin{figure}%[!h]
	\hspace{0.2cm}
	\center
	\includegraphics[width=0.92 \columnwidth]{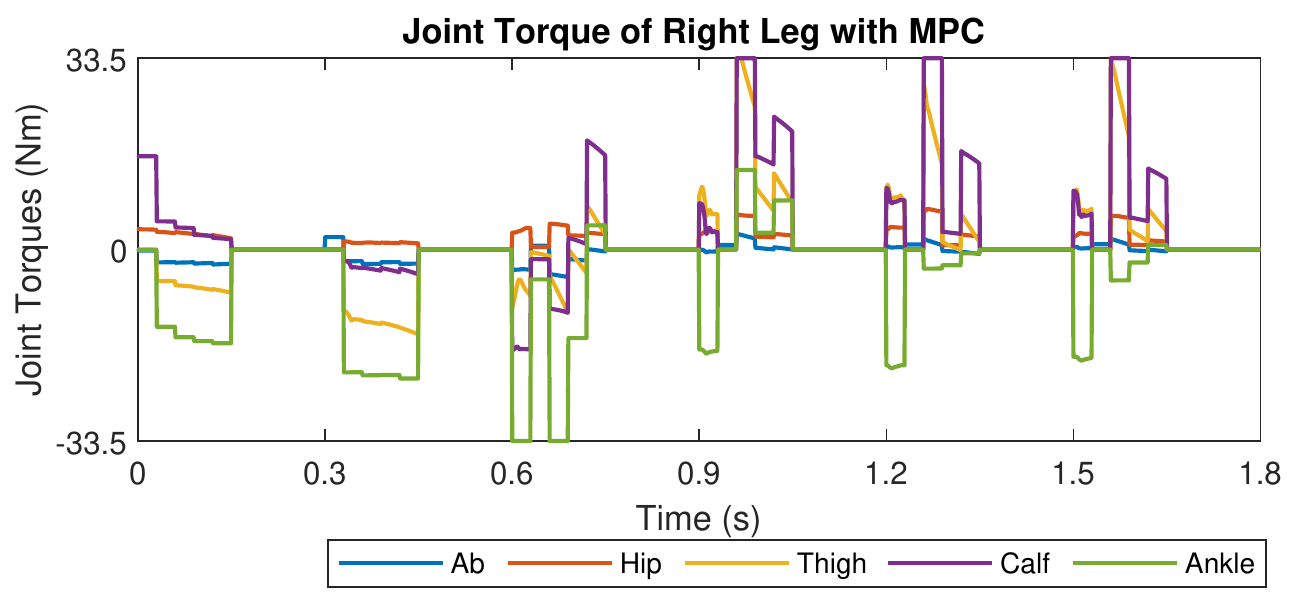}
	\caption{{\bfseries Plots of Joint Torques in Rough Terrain Simulation.  } }
	\label{fig:MPCtauRT}
\end{figure}

Plots of CoM location, velocity, and body orientation are shown in Fig. \ref{fig:CoMposVel} and Fig. \ref{fig:CoMeulAng}. It can be observed that the CoM location and orientation during this simulation maintain small tracking errors. 
%The position curves of the joints show that the joint position responses are smooth under the control of MPC and PD Cartesian control. The MPC controller input is presented in Fig. \ref{fig:MPCforce}. 
%The force in the $y$-direction $\bm F_y$ and moment $\bm M_z$ has the largest variation between the two legs. Hence it yields a slight $y$-direction displacement at the end of the simulation. As can be seen in CoM $y$-direction location in Fig. \ref{fig:CoMposVel}, the final $y$-direction location of body CoM is $0.0087 \:\unit m$ at $t = 1.8\: \unit{s}$. 
The joint torques (shown in Fig. \ref{fig:MPCtauRT}) during this entire simulation are in reasonable ranges and satisfy the torque saturation shown in Table \ref{tab:motor}. 
% todo{You should cite the table II about the torque limit here and also mention about the satisfaction of joint speed limit in Fig 11.} 
% (this is not really accurate so I commented it out)It is expected that the ankle joints will exert higher magnitudes of torques. Shown in the corresponding plots, the magnitudes of torques are still under the maximum torque threshold in most occasions. 

%With above simulation results and observations. It is inferred that the framework with the new MPC model presented in this paper can be a feasible option for a 10 DoF bipedal robot in dynamic locomotion. Our future work include extending this control framework to more dynamic motions such as bipedal bounding and running \todo{We have new results so please check sth like this throughout the paper to make sure that it is consistent with the current result}. Eventually, this control framework is expected to be migrated to a physical bipedal robot platform that is under development currently.
%\begin{figure}
%	\hspace{0.2cm}
%	\center
%	\includegraphics[width=1 \columnwidth]{Figures/RTJoint_final.pdf}
%	\caption{{\bfseries Plots of Joint Position and Velocity  in Rough Terrain Simulation. }  \todo{Remove this}}
%	\label{fig:jointPos}
%\end{figure}
\begin{figure}[!h]
	\center
	\includegraphics[width=0.92 \columnwidth]{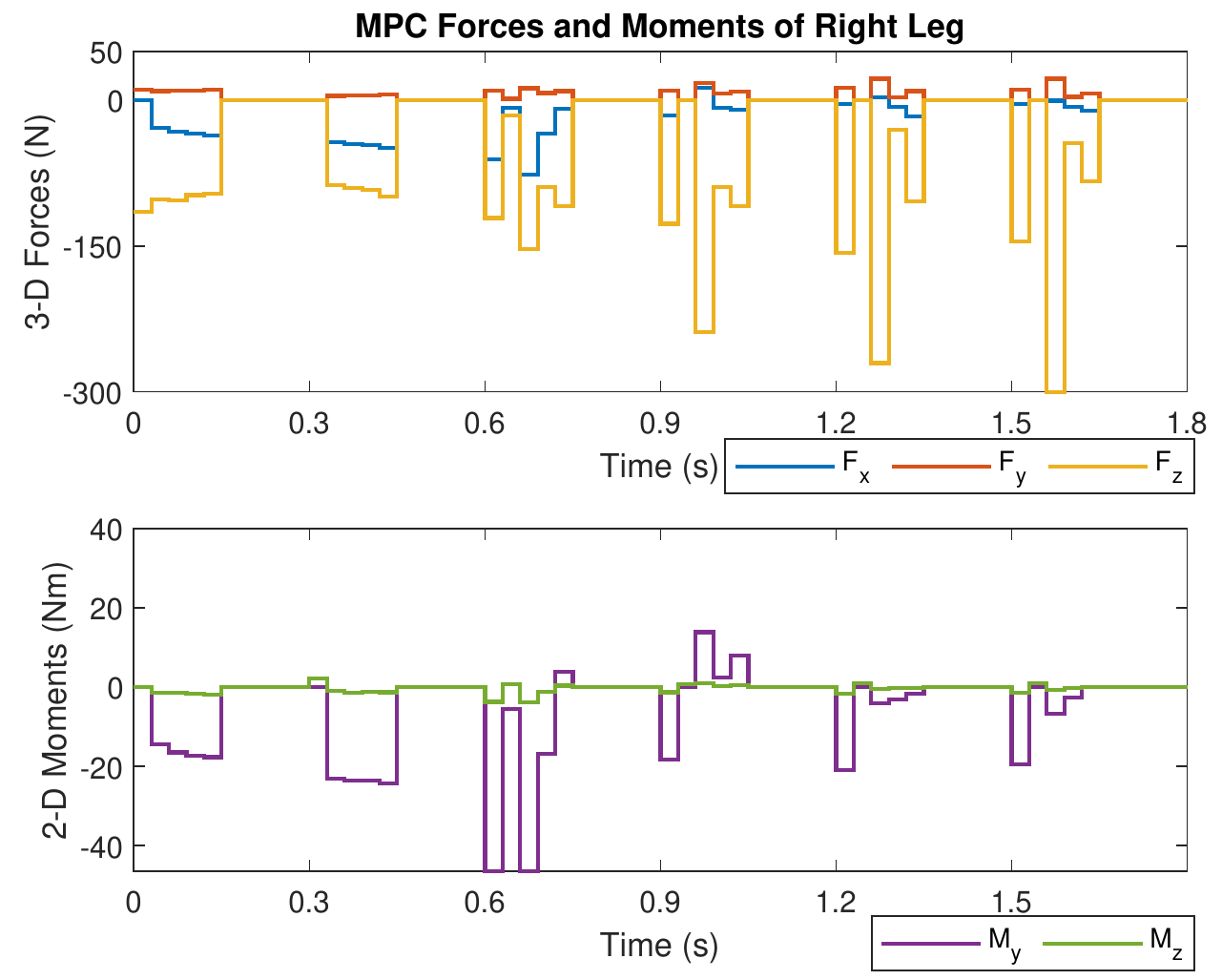}
	\caption{{\bfseries Plots of MPC Force and Moment in Rough Terrain Simulation.  }}
	\label{fig:MPCforce}
\end{figure}

\subsection{Bipedal Hopping}
\begin{figure}%[!h]
	\center
	\includegraphics[width=1 \columnwidth]{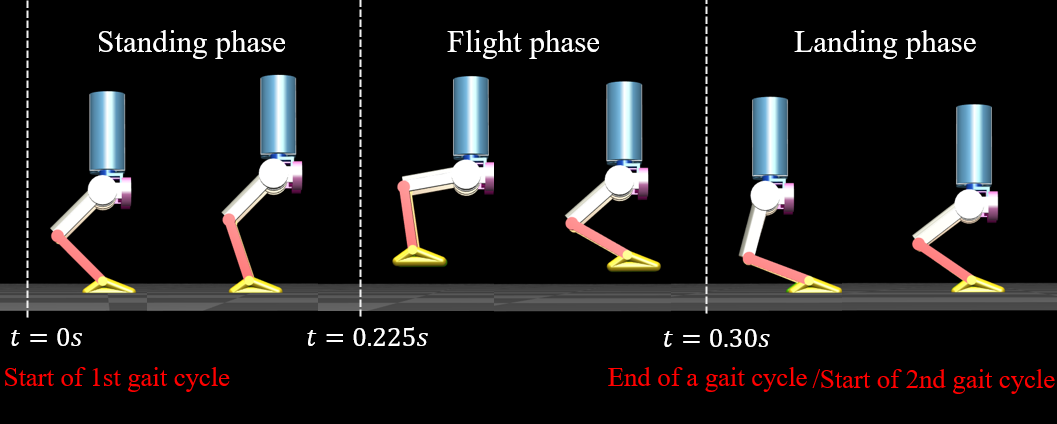}
	\caption{{\bfseries Illustration of Bipedal Hopping in Simulation  }  }
	\label{fig:boundGait}
\end{figure}

On top of the rotation and walking simulations presented earlier in this section, we have also implemented other gaits such as hopping. The hopping gait consists of a double support phase and a flight phase during the last quarter of each gait. 
% (we do not have flight phase in the previous one with trotting, so we should not mention this)With current MPC formulation, we decrease the flight phase duration in each gait cycle to mitigate the effects on performance during flight phase. 
A hopping gait illustration is shown in Fig. \ref{fig:boundGait}. It can be observed that during hopping motion, the robot is in a clear flight phase. 
%To validate the feasibility of hopping motion with the current MPC formulation, we test hopping forward motion with velocity $\bm v_{x_d}=0.5\:\unit{m/s}$. The CoM z-direction position and velocity plots are shown in Fig. \ref{fig:bounding_CoMposvel}. It is observed that the hopping motion can be performed with the current MPC formulation, with the trade-off of a certain degree of error in CoM velocity and height tracking. 
\update{This result validated that our proposed approach can work effectively for different dynamic locomotion on bipedal robots. We plan to optimize the MPC formulation in future work to enable faster and more aggressive motions. }

%\begin{figure}[!h]
%	\hspace{0.2cm}
%	\center
%	\includegraphics[width=1 \columnwidth]{Figures/BoundingCoM.pdf}
%	\caption{{\bfseries Plots of Z-direction CoM Location and Velocity in Hopping Simulation }  }
%	\label{fig:bounding_CoMposvel}
%\end{figure}

%% file: Sections/Conclusion.tex
%!TEX root = ../Main.tex

\section{Conclusions}
\label{sec:Conclusion}
In conclusion, we introduced an effective approach of force-and-moment-based Model Predictive Control to achieve highly dynamic locomotion on rough terrains for 10 degrees of freedom bipedal robots. Our framework also allows the robot to achieve a wide range of 3-D motions using the same control framework with the same set of control parameters. 
% This control framework and robot model is designed to perform in highly dynamic 3-D motion. 
The convex MPC formulation can be translated into a Quadratic Program problem and solved effectively in real-time of less than $1 \unit{ms}$. 
% Working in conjunction with Cartesian PD control for swing leg, we validate the MPC performance in a simulation. 
We explore and find the most suitable dynamics model for the control framework and we have presented successful walking simulations with time-varying velocity input, rough-terrain locomotion with high velocity and results in different dynamic gaits. Simulation results have indicated that the control performance in the velocity tracking test has a maximum deviation of $0.076\: \unit{m/s} $ compared to the desired input. In the rough terrain test, the robot is able to walk through rough terrain with various heights while maintaining a high forward walking velocity at $1.6\:\unit{m/s}$. 
% The control framework can be extended to many other dynamic motions such as bipedal running and hopping. We are working on improving performance and modifying MPC formulation for each type of motion, to achieve better results. 
Future work will include extending the approach for more aggressive motion and experimental validation of the framework on the robot hardware.
% This control framework and MPC design are also expected to be extended to a physical bipedal robot for testing and validation in the near future. 